\documentclass[a4paper]{article}
\usepackage{graphicx}
\usepackage{amsmath}
\usepackage{amssymb}
\usepackage{amsthm}
\usepackage{enumerate}
\usepackage{url}
\usepackage{subfigure}
\usepackage{times}
\usepackage{color}
\usepackage{multirow}
\usepackage{bm}
\usepackage{xr}
\usepackage{url}
\usepackage{natbib}
\usepackage{subfigmat}

\usepackage{epstopdf}

\newcommand{\KDE}{\text{KDE}}

\newcommand{\LOOCV}{\mathrm{LOOCV}}

\newcommand{\K}{\bm{K}}
\newcommand{\G}{\bm{G}}
\renewcommand{\H}{\bm{H}}

\newcommand{\dx}{d_{\mathrm{x}}}

\newcommand{\D}{\mathcal{D}}

\newcommand{\kx}{k_{\mathrm{x}}}
\newcommand{\ky}{k_{\mathrm{y}}}

\newcommand{\intd}{\mathrm{d}}
\newcommand{\parder}[1]{\frac{\partial}{\partial #1}}
\newcommand{\calD}{\mathcal{D}}

\newcommand{\rkhs}{\mathcal{H}}

\newtheorem{theorem}{Theorem}

\newcommand{\I}{\mathbf{I}}
\newcommand{\A}{\mathbf{A}}
\newcommand{\te}{\mathrm{te}}

\newcommand{\fnn}{f_{\mathrm{NN}}}
\newcommand{\rnn}{r_{\mathrm{NN}}}

\def\<{\langle}
\def\>{\rangle}

\newcommand{\R}[1]{\mathbb{R}^{#1}}

\usepackage{newfloat}
\DeclareFloatingEnvironment[fileext=frm,placement=h!,name=Frame]{myfloat}
\newenvironment{myalgo}[1][0.45\textwidth]{
  \begin{myfloat}[h]
    \begin{flushleft}
      \rule{\linewidth}{0.5mm}
    \end{flushleft}
    \begin{flushleft}}
    {\end{flushleft}
    \begin{flushleft}
    \rule{\linewidth}{0.5mm}
    \end{flushleft}
  \end{myfloat}
}

\newcommand{\argmax}{\mathop{\rm argmax}\limits}
\newcommand{\argmin}{\mathop{\rm argmin}\limits}

\newcommand{\wfnn}{\widehat{f}_{\mathrm{NN}}}
\newcommand{\wrnn}{\widehat{r}_{\mathrm{NN}}}
\newcommand{\km}{k_{\mathrm{m}}}
\newcommand{\bkm}{\bm{k}_{\mathrm{m}}}

\title{Robust modal regression with direct log-density
derivative estimation}
\author{Hiroaki Sasaki\\
Department of Complex and Intelligent Systems\\
Future University Hakodate, Japan
\\\vspace{-1mm} \\
Tomoya Sakai \\
NEC Corporation, Japan
\\\vspace{-1mm} \\
Takafumi Kanamori\\
Department of Mathematical and Computing Science\\
Tokyo Institute of Technology, Japan\\
RIKEN AIP, Japan
}
\date{}
\begin{document}
\maketitle
\begin{abstract} 
 Modal regression is aimed at estimating the global mode (i.e., global
 maximum) of the conditional density function of the output variable
 given input variables, and has led to regression methods robust against
 heavy-tailed or skewed noises. The conditional mode is often estimated
 through maximization of the modal regression risk (MRR). In order to
 apply a gradient method for the maximization, the fundamental challenge
 is accurate approximation of the gradient of MRR, not MRR itself. To
 overcome this challenge, in this paper, we take a novel approach of
 directly approximating the gradient of MRR. To approximate the
 gradient, we develop kernelized and neural-network-based versions of
 the least-squares log-density derivative estimator, which directly
 approximates the derivative of the log-density without density
 estimation.  With direct approximation of the MRR gradient, we first
 propose a modal regression method with kernels, and derive a new
 parameter update rule based on a fixed-point method. Then, the derived
 update rule is theoretically proved to have a monotonic hill-climbing
 property towards the conditional mode.  Furthermore, we indicate that
 our approach of directly approximating the gradient is compatible with
 recent sophisticated stochastic gradient methods (e.g., Adam), and then
 propose another modal regression method based on neural
 networks. Finally, the superior performance of the proposed methods is
 demonstrated on various artificial and benchmark datasets.
\end{abstract}
 \section{Introduction}
 \label{sec:intro}
 Recently, modal regression has been gathering a great deal of attention
 due to the clear advantages over conventional regression methods based
 on the conditional
 mean~\citep{sager1982maximum,collomb1986note,carreira2000reconstruction,einbeck2006modelling,yao2012local,chen2016nonparametric,feng2017statistical,wang2017regularized}.
 Modal regression can be roughly divided into \emph{unimodal} and
 \emph{multimodal regression}. The goal of the unimodal regression is to
 estimate the global mode (i.e., global maximum) of the conditional
 density, leading to regression methods robust against skewed or
 heavy-tailed
 noises~\citep{sager1982maximum,collomb1986note,yao2012local,feng2017statistical},
 while the conventional conditional mean estimation could be vulnerable
 to these nonGaussian noises.  On the other hand, multimodal regression
 is aimed at estimating local modes (i.e., local maxima) of the
 conditional density, and simultaneously finds multiple functional
 relationships between input and output variables which the conditional
 mean cannot
 capture~\citep{carreira2000reconstruction,einbeck2006modelling,chen2016nonparametric,sasaki2016modal}.
 Modal regression has been applied to a wide-range of research fields
 such as prediction of Alzheimer's disease~\citep{wang2017regularized},
 analysis of speed-flow data~\citep{einbeck2006modelling}, face
 recognition~\citep{8672460}, etc.~(See also a recent comprehensive
 review article by~\citet{chen2018modal}.) The scope of this paper is
 unimodal regression, which is called simply modal regression in the
 rest of this paper.
  
 The mode of the conditional density has been often estimated through
 maximization of the empirical \emph{modal regression risk} (MRR), which
 is defined as the sample average of the conditional density (or the
 joint
 density)~\citep{sager1982maximum,yao2012local,feng2017statistical}. A
 naive approach in modal regression takes a two-step approach of firstly
 approximating the empirical MRR via conditional (or joint) density
 estimation (e.g., by kernel density estimation), and secondly of
 maximizing the approximated risk by some gradient method. However, the
 crucial quantity in maximization is the gradient of the empirical MRR
 rather than MRR itself. Thus, this two-step approach might be
 suboptimal because a good MRR approximator does not necessarily mean a
 good gradient approximator of MRR.  \citet{yao2012local} employed an EM
 algorithm, but still computes the zero-crossing of the gradient
 obtained through the two-step approach. 

 Another approach in modal regression employs a surrogate risk of
 MRR~\citep{lee1989mode,yao2014new,feng2017statistical,wang2017regularized}. The
 advantage of this approach is that high-dimensional density estimation
 can be avoided. However, a drawback is that the surrogate risk includes
 a manually tuning hyperparameter, and it is not straightforward to
 select it since the surrogate risk itself depends on the
 hyperparameter. Moreover, when neural networks are employed for large
 scale datasets, the hyperparameter selection only can be computationally
 expensive.

 In this paper, we propose two methods for modal regression based on
 reproducing kernels and neural networks, respectively. In stark contrast with
 existing methods, we do not go through the approximation of MRR itself,
 but rather more directly approximate the gradient of MRR. The key
 challenge in the direct approximation is accurate estimation of
 (log-)density derivatives. To this end, we employ the Fisher divergence
 and develop a direct estimator for log-density derivatives without
 resorting to density
 estimation~\citep{cox1985penalty,SasakiHS14clustering}.
 
 First, we develop a modal regression method based on reproducing
 kernels. As shown later, thanks to the analytic solution of our
 log-density derivative estimator, a computationally efficient model
 selection is possible for leave-one-out cross validation. Furthermore,
 in modal regression, this kernel-based log-density derivative estimator
 enables to derive a novel parameter update rule based on a fixed-point
 method for conditional mode estimation, and we theoretically prove that
 the derived parameter update rule has a monotonic hill climbing
 property under some conditions.
 
 Next, we propose a modal regression method based on neural
 networks. The challenge is to stochastically estimate the conditional
 mode. Our approach of directly estimating the gradient of MRR is rather
 compatible with recent sophisticated stochastic gradient methods: The
 learning rates in AdaGrad~\citep{duchi2011adaptive},
 RMSprop~\citep{Hinton2012} and Adam~\citep{kingma2015adam} are
 adaptively determined by the gradient of an empirical risk. Thus,
 combined with these stochastic gradient methods, we can develop a
 neural-network-based method in a straightforward way, and to the best
 of our knowledge, this is the first attempt to make use of neural
 networks in modal regression. Finally, we demonstrate that our
 regression methods with reproducing kernels and neural networks work
 well on various artificial and benchmark datasets.
 \section{Background}
 This section gives some background of modal regression and states our
 approach.
  \subsection{Problem formulation}
  Suppose that we are given $n$ observations of pairs of input and
  output variables drawn from the joint density $p(y,\bm{x})$ for
  $y\in\R{}$ and $\bm{x}\in\R{\dx}$ as
  \begin{align*}
   \calD:=\left\{(y_i, \bm{x}_i^{\top})^{\top}\right\}_{i=1}^n.
  \end{align*}
  Under the assumption that the global conditional mode uniquely exists,
  our goal is to estimate the following \emph{modal regression function}
  $f_{\mathrm{M}}$ from $\calD$:
  \begin{align}
   f_{\mathrm{M}}(\bm{x}):=\argmax_{t\in\R{}} \log p(t|\bm{x}).
   \label{modal-func}
  \end{align}
  \subsection{Review of modal regression}
  \label{ssec:review}
  To make our approach clearer, we adopt the terminologies
  in~\citet{feng2017statistical}. Let us assume that the output variable
  $y$ is generated from the following model:
  \begin{align}
   y=f^*(\bm{x})+\epsilon(\bm{x}), \label{regression}
  \end{align}
  where $f^*$ and $\epsilon$ denote an unknown function and an additive
  noise, respectively. In~\eqref{regression}, the fundamental assumption
  is that the global mode of the conditional probability density
  function of $\epsilon$ given $\bm{x}$ is zero. This conditional mode
  assumption is much weaker than the standard Gaussian noise assumption
  because the noise $\epsilon$ can be skewed or heavy-tailed, or even
  have a nonstationary variance. The zero mode assumption ensures that
  $f_{\mathrm{M}}(\bm{x})=f^*(\bm{x})$. In order to have regression
  methods tolerable to heavy-tailed noises, another approach is to use
  robust loss functions~\citep{huber2009robust}, but is often intended
  for the (robustified) conditional mean estimation and thus might be
  vulnerable to skewed noises.
  
  To estimate $f_{\mathrm{M}}$ by a model $f_{\bm{\theta}}$ with
  parameters $\bm{\theta}$, the \emph{modal regression risk}
  (MRR)~\citep{feng2017statistical} is defined as
  \begin{align}
   \mathcal{R}(\bm{\theta}):= \int p(\bm{x})\log 
   p(f_{\bm{\theta}}(\bm{x})|\bm{x}) \intd \bm{x}.
   \label{modal-risk}
  \end{align}    
  An alternative risk has been also defined using the joint density
  $p(y,\bm{x})$~\citep{sager1982maximum,yao2012local} because
  $f_{\mathrm{M}}(\bm{x})=\arg\max_{t\in\R{}} p(t,\bm{x})$
  from~\eqref{modal-func}. Following Theorem~3
  in~\citet{feng2017statistical}, it can be proved that the (global)
  maximizer of $\mathcal{R}(\bm{\theta})$ equals to $f_M$ when both
  $f_{\bm{\theta}}$ and $f_M$ belong to the same function set. In
  practice, the empirical version of $\mathcal{R}(\bm{\theta})$ is used
  as
  \begin{align}
   \widetilde{\mathcal{R}}(\bm{\theta}):=\frac{1}{n}\sum_{i=1}^n \log
   p(f_{\bm{\theta}}(\bm{x}_i)|\bm{x}_i).  \label{empirical-modal-risk}
  \end{align}
  Then, $\widetilde{\mathcal{R}}(\bm{\theta})$ can be maximized based on
  the following gradient with respect to parameters $\bm{\theta}$:
  \begin{align}
   \parder{\bm{\theta}}\widetilde{\mathcal{R}}(\bm{\theta})
   &=\frac{1}{n}\sum_{i=1}^n
   \left\{\parder{\bm{\theta}} f_{\bm{\theta}}(\bm{x}_i)\right\}
   \parder{y}\log p(y|\bm{x}_i)\Bigr|_{y=f_{\bm{\theta}}(\bm{x}_i)}
   \nonumber\\
   &=\frac{1}{n}\sum_{i=1}^n
   \left\{\parder{\bm{\theta}} f_{\bm{\theta}}(\bm{x}_i)\right\}
   \parder{y}\log p(y,\bm{x}_i)\Bigr|_{y=f_{\bm{\theta}}(\bm{x}_i)},
   \label{grad-risk}
  \end{align}
  where note that $\parder{y}\log p(y|\bm{x})=\parder{y}\log
  p(y,\bm{x})$. To approximate the gradient~\eqref{grad-risk}, we need
  to estimate $\parder{y}\log p(y|\bm{x})$ or $\parder{y}\log
  p(y,\bm{x})$. To estimate the log-density derivative, a naive approach
  takes two steps of firstly estimating $\log p(y|\bm{x})$ or $\log
  p(y,\bm{x})$ and then of computing the derivative with respect to
  $y$. However, such a naive estimation procedure can be suboptimal
  because a good density estimator does not necessarily mean a good
  log-density derivative estimator.  Thus, a more reliable approach to
  approximate the gradient~\eqref{grad-risk} would be to directly
  estimate the log-density derivative $\parder{y}\log p(y|\bm{x})$ or
  $\parder{y}\log p(y,\bm{x})$ without going through density estimation.
    
  Another approach employs the following empirical surrogate
  risk~\citep{lee1989mode,yao2014new,feng2017statistical,wang2017regularized},
  which has been also used in the maximum correntropy
  criterion~\citep{gunduz2009correntropy,he2010maximum,feng2015learning}:
  \begin{align}
   \widetilde{\mathcal{R}}^{\sigma}(\bm{\theta}):=\frac{1}{n\sigma}
   \sum_{i=1}^n
   \psi\left(\frac{y_i-f_{\bm{\theta}}(\bm{x}_i)}{\sigma}\right),
   \label{surrogate}
  \end{align}
  where $\sigma$ is a positive width parameter, $\psi$ is a nonnegative
  function such that $\psi(u)=\psi(-u)$, $\psi(u)\leq \psi(0)$ for all
  $u$ and $\int \psi(u)\intd u=1$. \citet{feng2017statistical} proved
  the following relation:
  \begin{align*}
   \widetilde{\mathcal{R}}^{\sigma}(\bm{\theta})
   &\xrightarrow[]{n\rightarrow\infty} \frac{1}{\sigma}\int
   \psi\left(\frac{y-f_{\bm{\theta}}(\bm{x})}{\sigma}\right)
   p(y,\bm{x})\intd y\intd\bm{x}\xrightarrow[]{\sigma\rightarrow0}
   \int p(f_{\bm{\theta}}(\bm{x})|\bm{x}) p(\bm{x})\intd\bm{x}.
  \end{align*}
  Thus, $\widetilde{\mathcal{R}}^{\sigma}(\bm{\theta})$ can be regarded
  as a surrogate of~$\widetilde{\mathcal{R}}(\bm{\theta})$
  in~\eqref{empirical-modal-risk} without the logarithm. This approach
  seems appealing because we can avoid high-dimensional density
  estimation. On the other hand, a significant drawback is that the
  performance strongly depends on the choice of the hyperparameter
  $\sigma$, and it is not straightforward to choose a right value. We
  may use cross validation (CV) in practice, but this approach can be
  problematic because of the following two reasons: First, it seems
  unclear what criterion in CV should be used to select $\sigma$ because
  $\widetilde{\mathcal{R}}^{\sigma}$ itself depends on
  $\sigma$\footnote{The squared-loss may be used in CV. However, the
  squared-loss implicitly assumes the Gaussian noise, and thus may
  prohibit us to make full use of the advantages of modal regression.};
  Second, even if there was a valid criterion for CV, then we have to
  perform a nested CV to choose both $\sigma$ and hyperparameters in
  $f_{\bm{\theta}}$ (e.g., the width parameter in a kernel function),
  which tends to be computationally very expensive. Furthermore, if
  neural networks are employed, a grid-search of $\sigma$ only could be
  computationally costly in high-dimensional and large datasets.

  Here, our approach is to directly approximate the gradient
  $\parder{\bm{\theta}}\widetilde{\mathcal{R}}(\bm{\theta})$ without any
  approximation of the empirical modal regression risk
  $\widetilde{\mathcal{R}}(\bm{\theta})$ itself.  To this end, the key
  idea is to directly estimate the log-density derivative
  $\parder{y}\log p(y,\bm{x})$ in~\eqref{grad-risk}. With the direct
  approximation, we propose two novel methods for modal regression using
  reproducing kernels and neural networks.
 \section{Direct log-density derivative estimation with reproducing kernels} 
 \label{sec:K-LSDDR}
 This section adopts a direct approach for log-density derivative
 estimation~\citep{cox1985penalty,SasakiHS14clustering}, and derives an
 estimator based on reproducing kernels. Here, our contributions are to
 establish Theorem~\ref{theo:analytic-solution} and to show an analytic
 form of the leave-one-out cross-validation score for model selection.
 %
 %  %
  \subsection{Kernelized estimator for log-density derivatives}
  To estimate the log-density derivative, we directly fit a model
  $r(y,\bm{x})$ under the Fisher
  divergence~\citep{cox1985penalty,SasakiHS14clustering}:
  \begin{align}
   J(r)&:=\frac{1}{2}\int\{r(y,\bm{x})-\parder{y}\log
   p(y,\bm{x})\}^2 p(y,\bm{x})\intd y\intd\bm{x}\nonumber\\
   &=\frac{1}{2}\int \{r(y,\bm{x})\}^2 p(y,\bm{x})\intd
   y\intd\bm{x} -\int
   r(y,\bm{x})\left\{\parder{y}p(y,\bm{x})\right\}\intd y\intd\bm{x}
   +\frac{1}{2}\int\left\{\parder{y}\log p(y,\bm{x})\right\}^2
   p(y,\bm{x})\intd y\intd\bm{x}.
   \label{Fisher-div}
  \end{align}
  Next, we apply the well-known
  \emph{integration by parts} technique to the second term
  in~\eqref{Fisher-div} as follows:
  \begin{align*}
   \int r(y,\bm{x})\left\{\parder{y}p(y,\bm{x})\right\}\intd y\intd\bm{x}
   =-\int \left\{\parder{y}r(y,\bm{x})\right\}p(y,\bm{x})\intd y\intd\bm{x},
  \end{align*}
  where we assumed that for all $\bm{x}$,
  \begin{align}
   \lim_{|y|\rightarrow\infty}r(y,\bm{x})p(y,\bm{x})=0.
   \label{K-LSLD-assumption}
  \end{align}
  Then, the empirical Fisher divergence up to the ignorable constant is
  obtained as
  \begin{align}
   \widehat{J}(r)&=\frac{1}{n}\sum_{i=1}^n\left[
   \frac{1}{2}r(y_i,\bm{x}_i)^2
   +\parder{y}r(y_i,\bm{x}_i)\right].
   \label{eqn:empirical-risk}
  \end{align} 
  
  Based on the empirical Fisher divergence, \citet{cox1985penalty}
  proposed a practical estimator with a one-dimensional piecewise
  polynomial kernel~\citep{wahba1990spline}, while
  \citet{SasakiHS14clustering} applied the $\ell_2$ regularizer for
  model parameters in $r(y,\bm{x})$.  Here, we employ the general
  kernel function and regularizer for RKHS norm, and our estimator is
  defined as
  \begin{align}
   \widehat{r}=\argmin_{r\in\rkhs}\left[
  \widehat{J}(r)+\frac{\lambda}{2}\|r\|_{\rkhs}^2
  \right],\label{eqn:K-LSDDR}
  \end{align}
  where $\|\cdot\|_{\rkhs}$ and $\lambda(>0)$ denote RKHS norm and the
  regularization parameter, respectively.  Then, the following theorem
  shows that $\widehat{r}$ can be efficiently obtained by solving
  systems of linear equations:
  \begin{theorem}
   \label{theo:analytic-solution} Let us express
   $(y,\bm{x}^{\top})^{\top}$ by $\bm{z}$. $\widehat{r}$ is given by
   \begin{align}
    \widehat{r}(\bm{z})&=\sum_{i=1}^n\left[\widehat{\alpha}_{i}
    k(\bm{z},\bm{z}_i)-\frac{1}{n\lambda}
    \parder{y'}k(\bm{z},\bm{z}')\Bigr|_{\bm{z}'=\bm{z}_i}\right],
   \label{optimal-estimator}
   \end{align}
   where $k(\bm{z},\bm{z}')$ denotes the kernel function,
   $\bm{z}_i:=(y_i,\bm{x}_i^{\top})^{\top}$ and
   $\bm{z}':=(y',\bm{x}^{'\top})^{\top}$.  The coefficients
   $\widehat{\bm{\alpha}}
   =(\widehat{\alpha}_1,\widehat{\alpha}_2,\dots,\widehat{\alpha}_{n})^{\top}$
   are the solution of the following system of linear equations:
   \begin{align}
    (\K+n\lambda\I_n)\widehat{\bm{\alpha}}
    =\frac{1}{n\lambda}\G\bm{1}_n,
    \label{solution-alpha}
   \end{align}
   where $\bm{1}_n=(1,1,\dots,1)^{\top}$ is an $n$-dimensional vector,
   $\I_n$ denotes the $n$ by $n$ identity matrix,
   $[\K]_{ij}=k(\bm{z}_i,\bm{z}_j)$ and
   $[\G]_{ij}=\parder{y'}k(\bm{z}_i,\bm{z}')|_{\bm{z}'=\bm{z}_i}$.
  \end{theorem}
  The proof is deferred to
  Appendix~\ref{app:proof-analytic-solution}. This paper calls this
  method the \emph{kernelized least-squares log-density derivatives}
  (K-LSLD).  Section~\ref{sec:MR} develops a modal regression method
  based on K-LSLD.
  \subsection{Leave-one-out cross-validation}
  \label{ssec:LOOCV}
  The performance of K-LSLD depends on model selection (parameters in
  the kernel function and regularization parameter). Here, we perform
  the leave-one-out cross-validation (LOOCV) for model selection whose
  score is given by
  \begin{align*}
   &\LOOCV =\frac{1}{n}\sum_{l=1}^n\left[\frac{1}{2}
   \{\widehat{r}^{(l)}(y_l,\bm{x}_l)\}^2
   + \parder{y}\widehat{r}^{(l)}(y_l,\bm{x}_l)\right],
  \end{align*}
  where $\widehat{r}^{(l)}$ denotes the estimator obtained from the
  collection of data samples except for the $l$-th data sample
  (i.e. $\calD\setminus(y_l,\bm{x}_l^{\top})^{\top}$).  LOOCV is usually
  time-consuming. However, thanks to the analytic solution in Theorem~1,
  the LOOCV score can be efficiently computed. Details are presented in
  Appendix~\ref{app:LOOCV}.
 \section{Modal regression with direct log-density derivative estimation}
 \label{sec:MR}
 This section first develops a kernel-based method for modal
 regression. Based on K-LSLD, we derive a parameter update rule based on
 a fixed-point method. Then, the derived update rule is theoretically
 investigated. Finally, another novel modal regression method is also
 proposed based on neural networks.
  \subsection{Direct modal regression with kernels}
  \label{ssec:DMR-K}
   \subsubsection{Fixed-point-based  parameter update rule}
   Here, we employ a model $f_{\bm{\theta}}$ in an RKHS to estimate the
   conditional mode. Then, under the empirical modal regression
   risk~\eqref{empirical-modal-risk}, the representer
   theorem~\citep{kimeldorf1971some,scholkopf2001learning} suggests the
   optimal form of $f_{\bm{\theta}}$ as
   \begin{align}
    f_{\bm{\theta}}(\bm{x})=\sum_{k=1}^n
    \theta_{k}\km(\bm{x},\bm{x}_k)=\bm{\theta}^{\top}
    \bkm(\bm{x}),\label{eqn:regression-model}
   \end{align}
   where $\km(\bm{x}, \bm{x}_i)$ denotes a kernel function,
    $\bkm(\bm{x})=(\km(\bm{x},\bm{x}_1),\km(\bm{x},\bm{x}_2),
    \dots,\km(\bm{x},\bm{x}_n))^{\top}$, and
    $\bm{\theta}=(\theta_1,\theta_2,\dots,\theta_n)^{\top}$.  By
    substituting~\eqref{eqn:regression-model} into
    $\parder{\bm{\theta}}\widetilde{\mathcal{R}}(\bm{\theta})$, we have
    the gradient of the empirical MRR as
   \begin{align}
    \parder{\bm{\theta}}\widetilde{\mathcal{R}}(\bm{\theta})
    &=\frac{1}{n}\sum_{i=1}^n
    \parder{y}\log p(y,\bm{x}_i)\Bigr|_{y=\bm{\theta}^{\top}\bkm(\bm{x}_i)}   
    \bkm(\bm{x}_i).
    \label{empirical-modal-risk-grad}
   \end{align}
   To approximate~\eqref{empirical-modal-risk-grad}, we employ K-LSLD to
   estimate $\parder{y}\log p(y,\bm{x})$. After approximating the
   gradient $\parder{\bm{\theta}}\widetilde{\mathcal{R}}(\bm{\theta})$,
   a straightforward approach to estimate $\bm{\theta}$ would be to use
   gradient ascent.  Alternatively, we derive a simpler update rule for
   $\bm{\theta}$ based on a fixed-point method, which does not require
   any tuning parameters.

   Let us express the kernel function in K-LSLD as
   $k(\bm{z},\bm{z}')=\ky(y,y')\times\kx(\bm{x},\bm{x}')$ where both
   $\ky$ and $\kx$ are kernel functions.  Then, K-LSLD is given by
   \begin{align}
    &\widehat{r}(y,\bm{x})\nonumber\\ &:=\sum_{l=1}^n
    \left\{\widehat{\alpha}_l\ky(y,y_l)-
    \frac{1}{n\lambda}\parder{y'}\ky(y,y')\Bigr|_{y'=y_l}
    \right\}\kx(\bm{x},\bm{x}_l), \label{eqn:LSDDR-joint}
   \end{align}
   Next, we restrict the form of $\ky$ as
   \begin{align*}
    \ky(y,y')=\phi\left\{\frac{(y-y')^2}{2\sigma_{\mathrm{y}}^2}\right\},
   \end{align*}
    where $\sigma_{\mathrm{y}}(>0)$ denotes the width parameter, $\phi$
    is a convex, and monotonically non-increasing function. For
    instance, $\phi(t)=\exp(-t)$, $\ky(y,y')$ is the Gaussian kernel.
    Substituting $\widehat{r}(y,\bm{x})$ into $\parder{y}\log
    p(y,\bm{x})$ in~\eqref{empirical-modal-risk-grad} enables to
    approximate the gradient
    $\parder{\bm{\theta}}\widetilde{\mathcal{R}}(\bm{\theta})$ as
    \begin{align}
     \parder{\bm{\theta}}\widetilde{\mathcal{R}}(\bm{\theta})
     &\approx\frac{1}{n}\sum_{i=1}^n
     \widehat{r}(\bm{\theta}^{\top}\bkm(\bm{x}_i),\bm{x}_i)
     \bkm(\bm{x}_i) \nonumber\\ 
     &=\bm{h}(\bm{\theta})-\H(\bm{\theta})\bm{\theta}, 
     \label{eqn:modal-risk-LSDDR}
    \end{align}
    where with $\varphi(t):=-\frac{\intd}{\intd t}\phi(t)$,
    \begin{align}
     \H(\bm{\theta})&:=\frac{1}{n^2\lambda\sigma_{\mathrm{y}}^2}
     \sum_{i=1}^n\sum_{l=1}^n
     \varphi\left\{\frac{(\bm{\theta}^{\top}\bkm(\bm{x}_i)-y_l)^2}
     {2\sigma_{\mathrm{y}}^2}\right\}
     \kx(\bm{x}_i,\bm{x}_l)\bkm(\bm{x}_i)\bkm(\bm{x}_i)^{\top},
     \label{defi-H}\\
     \bm{h}(\bm{\theta})&:=\frac{1}{n}\sum_{i=1}^n\sum_{l=1}^n
     \left[\widehat{\alpha}_l
     \phi\left\{\frac{(\bm{\theta}^{\top}\bkm(\bm{x}_i)-y_l)^2}
     {2\sigma_{\mathrm{y}}^2}\right\}
     +\frac{y_l}{n\lambda\sigma_{\mathrm{y}}^2}
     \varphi\left\{\frac{(\bm{\theta}^{\top}\bkm(\bm{x}_i)-y_l)^2}
     {2\sigma_{\mathrm{y}}^2}\right\}\right]
     \kx(\bm{x}_i,\bm{x}_l)\bkm(\bm{x}_i).\label{defi-h}
    \end{align}
    Then, under the assumption that $\H(\bm{\theta})$ is invertible,
    setting the right-hand side in~\eqref{eqn:modal-risk-LSDDR} to zero
    gives the following iterative update rule based on a fixed-point
    method:
    \begin{align}
     \bm{\theta}^{\tau+1}=\H^{-1}(\bm{\theta}^{\tau})\bm{h}(\bm{\theta}^{\tau}),
     \label{fixed-point1}
    \end{align}
    where $\bm{\theta}^{\tau}$ denotes the $\tau$-th update of
    $\bm{\theta}$. Multiplying $\H^{-1}(\bm{\theta}^{\tau})$ to the both
    sides of~\eqref{eqn:modal-risk-LSDDR} and applying the update
    rule~\eqref{fixed-point1} yields the following relation:
    \begin{align}
     \bm{\theta}^{\tau+1} &\approx
    \bm{\theta}^{\tau}+\H^{-1}(\bm{\theta}^{\tau})
    \parder{\bm{\theta}}\widetilde{\mathcal{R}}(\bm{\theta})
    \Bigr|_{\bm{\theta}=\bm{\theta}^{\tau}}.  \label{approx-grad-ascent}
    \end{align}  
    Eq.\eqref{approx-grad-ascent} indicates that the update
    rule~\eqref{fixed-point1} approximately performs gradient ascent to
    maximize $\widetilde{\mathcal{R}}(\bm{\theta})$ when
    $\H(\bm{\theta})$ is positive definite. We more rigorously
    investigate a theoretical property of the update
    rule~\eqref{fixed-point1} below.
    
    An outline of our kernel-based algorithm called the \emph{direct
    modal regression with kernels} (DMR-K) is given in Algorithm~1. The
    important problem is how to determine the initial parameters
    $\bm{\theta}_0$ because the maximization of the modal regression
    risk may require to solve a non-convex optimization problem. As a
    remedy, we first perform some regression method based on the squared
    loss or absolute deviations, and use the estimated coefficient
    vector as $\bm{\theta}_0$. In addition, to ensure that
    $\H(\bm{\theta})$ is invertible, we may add a small constant to the
    diagonals of $\H(\bm{\theta})$ in practice.
    
    \begin{myalgo}
     {\bf Algorithm~1: Direct modal regression with kernels (DMR-K)}
     \vspace{1mm}
     \newline 
     {\sf {\bf Input:} Data $\{(y_i,\bm{x}_i)\}_{i=1}^n$, initial
     parameters $\bm{\theta}_0$
     \begin{enumerate}
      \item Estimate $\parder{y}\log p(y,\bm{x})$ as in
	    Theorem~\ref{theo:analytic-solution}.
	  
      \item Substitute $\bm{\theta}_0$ into
	    $f(\bm{x})=\bm{\theta}^{\top}\bkm(\bm{x})$, and repeat to
	    update $\bm{\theta}$ by~\eqref{fixed-point1} until some
	    convergence criterion is satisfied.
     \end{enumerate}
     {\bf Output:}
     $\widehat{f}(\bm{x}):=\widehat{\bm{\theta}}^{\top}\bkm(\bm{x})$
     with the optimized $\widehat{\bm{\theta}}$.}
    \end{myalgo}
    \subsubsection{Monotonic hill-climbing property of DMR-K}
    Here, we theoretically investigate DMR-K. In particular, we focus on
    the \emph{monotonic hill-climbing property} where for every $\tau$,
    the following inequality holds:
    \begin{align*}
     \widetilde{\mathcal{R}}(\bm{\theta}^{\tau+1})
     -\widetilde{\mathcal{R}}(\bm{\theta}^{\tau})> 0.
    \end{align*}    
    This inequality indicates that $\bm{\theta}$ is updated such that
    $\widetilde{\mathcal{R}}$ is monotonically increased.  However, it
    is not straightforward to investigate the monotonic hill-climbing
    property in our method because there is no approximation of the
    empirical risk~$\widetilde{\mathcal{R}}(\bm{\theta})$.
    
    To cope with this problem, we employ the formula of \emph{path
    integral}: Regarding the vector field
    $\parder{\bm{\theta}}\widetilde{\mathcal{R}}(\bm{\theta})$ and a
    differentiable curve $\bm{\theta}(t)$ from
    $\bm{\theta}(0)=\bm{\theta}_1$ to $\bm{\theta}(1)=\bm{\theta}_2$,
    the path integral is given by
    \begin{align}
     D[\bm{\theta}_2|\bm{\theta}_1]:=
     \int_{0}^1 \<\parder{\bm{\theta}}\widetilde{\mathcal{R}}(\bm{\theta}(t)),
     \dot{\bm{\theta}}(t)\>\intd t
     &=\widetilde{\mathcal{R}}(\bm{\theta}_2)
     -\widetilde{\mathcal{R}}(\bm{\theta}_1),
     \label{path-integral-UR}
    \end{align}
    where $\dot{\bm{\theta}}(t):=\frac{\intd}{\intd t}\bm{\theta}(t)$
    and $\<\cdot,\cdot\>$ denotes the inner product. The key point is
    that the right-hand side is independent to any choice of paths and
    computed only from $\bm{\theta}_1$ and $\bm{\theta}_2$. Our analysis
    uses the following simple path:
    \begin{align}
     \bm{\theta}(t)=\bm{\theta}_1+t(\bm{\theta}_2-\bm{\theta}_1),
     \label{linear-path}
    \end{align}
    where $0\leq t \leq 1$.
    
    Eq.\eqref{path-integral-UR} indicates that substituting our gradient
    approximator~\eqref{eqn:modal-risk-LSDDR} into
    $\parder{\bm{\theta}}\widetilde{\mathcal{R}}(\bm{\theta})$ in
    \eqref{path-integral-UR} gives us an approximator of
    $\widetilde{\mathcal{R}}(\bm{\theta}_2)-\widetilde{\mathcal{R}}(\bm{\theta}_1)$.
    Thus, we approximate the path integral
    $D[\bm{\theta}_2|\bm{\theta}_1]$ by our gradient
    approximator~\eqref{eqn:modal-risk-LSDDR} as
    \begin{align}
     &\widehat{D}[\bm{\theta}_2|\bm{\theta}_1] :=
     \frac{1}{n}\sum_{i=1}^n\int_0^1
     \widehat{r}(\bm{\theta}(t)^{\top}\bm{k}(\bm{x}_i),\bm{x}_i)
     \bm{k}(\bm{x}_i)^{\top}(\bm{\theta}_2-\bm{\theta}_1)\intd t,
     \label{path-integral-estimate-UR}
    \end{align}
    where the path~\eqref{linear-path} is applied.  When
    $\widehat{D}[\bm{\theta}^{\tau+1}|\bm{\theta}^{\tau}]> 0$ for every
    $\tau$, our update rule~\eqref{fixed-point1} can be regarded as
    having the monotonic hill-climbing property. The following theorem
    establishes sufficient conditions for the monotonic hill-climbing
    property:
    \begin{theorem}
     \label{theo:LSUR-rule} Assume that $\kx$ is non-negative, and
     $\phi$ is a convex, and monotonically non-increasing
     function. Then, if $\bm{\theta}^{\tau}\neq\bm{\theta}^{\tau+1}$,
     $\bm{H}(\bm{\theta})$ is positive definite and
     $\widehat{\alpha}_l=0$ for all $l$, under the update
     rule~\eqref{fixed-point1}, the following inequality holds:
     \begin{align*}
      \widehat{D}[\bm{\theta}^{\tau+1}|\bm{\theta}^{\tau}]> 0.
     \end{align*}
    \end{theorem}
    The proof is deferred to Appendix~\ref{app:LSUR-rule}. Conditions
    for $\kx$ and $\phi$ can be easily satisfied by using the Gaussian
    kernel, which also ensures that $\bm{H}(\bm{\theta})$ is positive
    definite by definition~\eqref{defi-H}. On the other hand, the
    condition for $\widehat{\alpha}_l$ is not satisfied in general.
    However, we experimentally observed that the update
    rule~\eqref{fixed-point1} gives good results without satisfying the
    condition $\widehat{\alpha}_l=0$. This would be because the update
    rule~\eqref{fixed-point1} possibly performs gradient ascent as
    implied in~\eqref{approx-grad-ascent}, and we conjecture that there
    exists milder conditions to improve Theorem~\ref{theo:LSUR-rule}.
    
    A similar analysis using path integral has been done in mode-seeking
    clustering~\citep{sasaki2018mode}. However, \citet{sasaki2018mode}
    proved a monotonic hill-climbing property with respect to the
    probability density function, while our analysis is for the
    empirical modal regression risk. Thus, the proof is substantially
    different.
  \subsection{Direct modal regression with neural networks}
  Here, we propose another modal regression method based on neural
  networks.  With a neural network $\fnn(\bm{x};\bm{\theta})$
  parametrized by $\bm{\theta}$, we directly compute the gradient of the
  empirical modal regression risk as follows:
  \begin{align}
   &\parder{\bm{\theta}}\widetilde{\mathcal{R}}(\bm{\theta})
   =\frac{1}{n}\sum_{i=1}^n \left\{\parder{\bm{\theta}}
   \fnn(\bm{x}_i;\bm{\theta})\right\}\parder{y}\log
   p(y,\bm{x}_i)\Bigr|_{y=\fnn(\bm{x}_i;\bm{\theta})}. \label{grad-risk-NN}
  \end{align}
  Our approach of directly approximating the gradient of an empirical
  risk~\eqref{grad-risk-NN} is rather well-compatible with recent
  sophisticated stochastic gradient methods: The learning rates are
  adaptively determined based on the gradient of an (mini-batch)
  empirical risk in AdaGrad~\citep{duchi2011adaptive},
  RMSprop~\citep{Hinton2012} and Adam~\citep{kingma2015adam}. Thus,
  estimating only the gradient still enables to use these stochastic
  optimization methods in a straightforward way.
 
  In addition to the conditional mode, we estimate $\parder{y}\log
  p(y,\bm{x})$ using a neural network model $\rnn(y,\bm{x};\bm{\gamma})$
  with parameters $\bm{\bm{\gamma}}$ based on the Fisher
  divergence. However, we experimentally observed that the second term
  in the empirical Fisher divergence~\eqref{eqn:empirical-risk} often
  diverged when feedforward neural networks were employed for
  $\rnn(y,\bm{x};\bm{\gamma})$. This is presumably because neural
  networks can be unbounded functions, and therefore it would be
  difficult to satisfy Assumption~\eqref{K-LSLD-assumption}.  To cope
  with this problem, we use the following form for
  $\rnn(y,\bm{x};\bm{\bm{\gamma}})$:
  \begin{align}
   \rnn(y,\bm{x};\bm{\gamma}) =\sum_{k=1}^K w_k
   \exp\left[-\frac{\{y-\mu_k^{\mathrm{NN}}(\bm{x})\}^2}{2\sigma_k^2}\right],
   \label{rnn-model}
  \end{align}
  where $w_k$ are parameters to be estimated, $\sigma_k$ denote (fixed)
  width parameters, and $\mu_k^{\mathrm{NN}}$ are modelled by neural
  networks. This model would satisfy
  Assumption~\eqref{K-LSLD-assumption} because $\rnn$ approaches to zero
  as $|y|\rightarrow\infty$.

  An outline of our algorithm called the \emph{direct modal regression
  with neural networks} (DMR-NN) is summarized in Algorithm~2.  As in
  DMR-K, it is an important problem to choose good initial parameters
  $\bm{\theta}_0$. Here, we perform \emph{pretraining} where
  $\fnn(\bm{x};\bm{\theta})$ is trained based on the squared loss or
  absolute deviations in advance.
  \begin{myalgo}
   {\bf Algorithm~2: Direct modal regression with neural networks (DMR-NN)}
   \vspace{1mm}
   \newline 
   {\sf {\bf Input:} Data $\{(y_i,\bm{x}_i)\}_{i=1}^n$, initial
   parameters $\bm{\theta}_0$
   \begin{enumerate}
    \item Estimate $\parder{y}\log p(y,\bm{x})$ by a
	  neural-network-based model $\rnn(y,\bm{x};\bm{\gamma})$
	  through minimization of the empirical Fisher
	  divergence~\eqref{eqn:empirical-risk} with a minibatch
	  stochastic gradient method.
	  
    \item Repeat the following with the log-density derivative estimator
	  $\wrnn(y,\bm{x})=\rnn(y,\bm{x};\widehat{\bm{\gamma}})$ of the
	  optimized $\widehat{\bm{\gamma}}$ and a neural network
	  $\fnn(\bm{x};\bm{\theta})$ initialized by
	  $\bm{\theta}=\bm{\theta}_0$:
	  \begin{enumerate}
	   \item With a random minibatch $\{\bm{x}^{(B)}_b\}_{b=1}^B$,
		 approximate the gradient~\eqref{grad-risk-NN} by
		 \begin{align*}		 
		  \bm{g}^{(B)}=\frac{1}{B}\sum_{b=1}^B		  
		  \left\{\parder{\bm{\theta}}\fnn(\bm{x}^{(B)}_b;\bm{\theta})\right\}
		  \wrnn(\fnn(\bm{x}^{(B)}_b;\bm{\theta}),\bm{x}^{(B)}_b). 
		 \end{align*}
		 
	   \item Update $\bm{\theta}$ by applying a minibatch stochastic
		 gradient method (e.g., Adam) using $\bm{g}^{(B)}$.
	  \end{enumerate}
   \end{enumerate}  
   {\bf Output:} $\wfnn(\bm{x}):=\fnn(\bm{x};\widehat{\bm{\theta}})$
   with the optimized $\widehat{\bm{\theta}}$}
  \end{myalgo}

  \section{Numerical illustration}
  Here, we numerically illustrate the performance of DMR-K and DMR-NN
  and compare them with existing methods.
  \subsection{Illustration of DMR-K on artificial datasets}
 \begin{figure}[t]
  \centering
  \subfigure{\includegraphics[width=.95\textwidth]{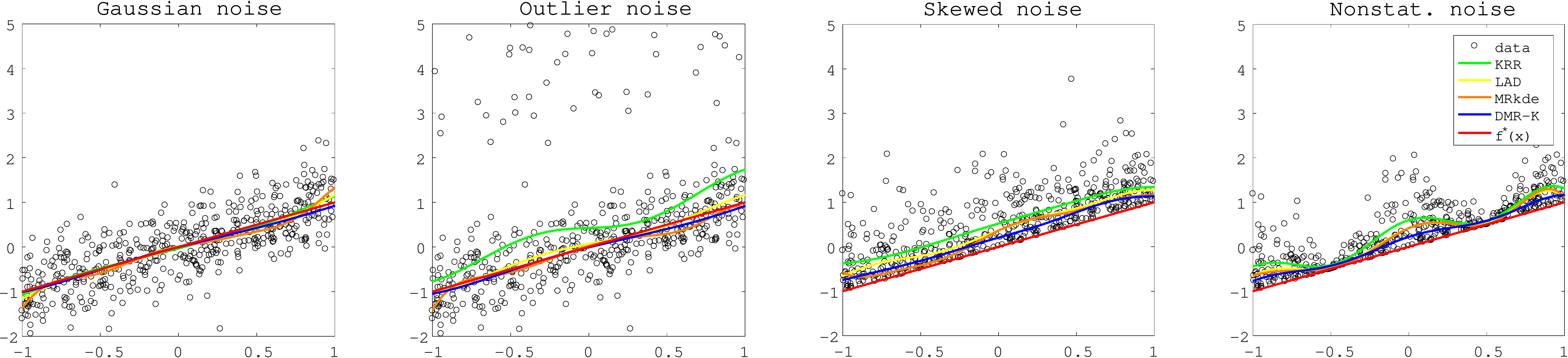}}
  \subfigure{\includegraphics[width=.95\textwidth]{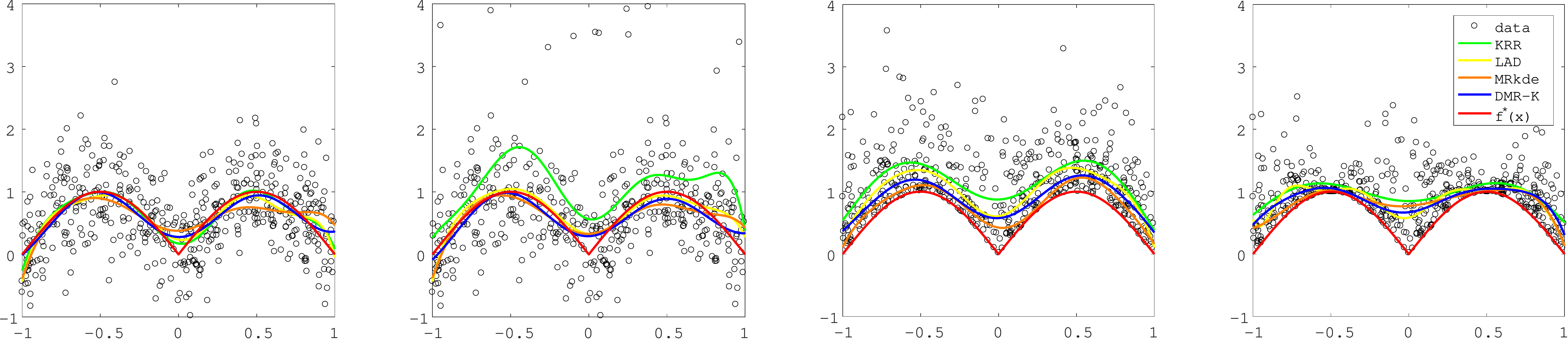}}
  \subfigure{\includegraphics[width=.95\textwidth]{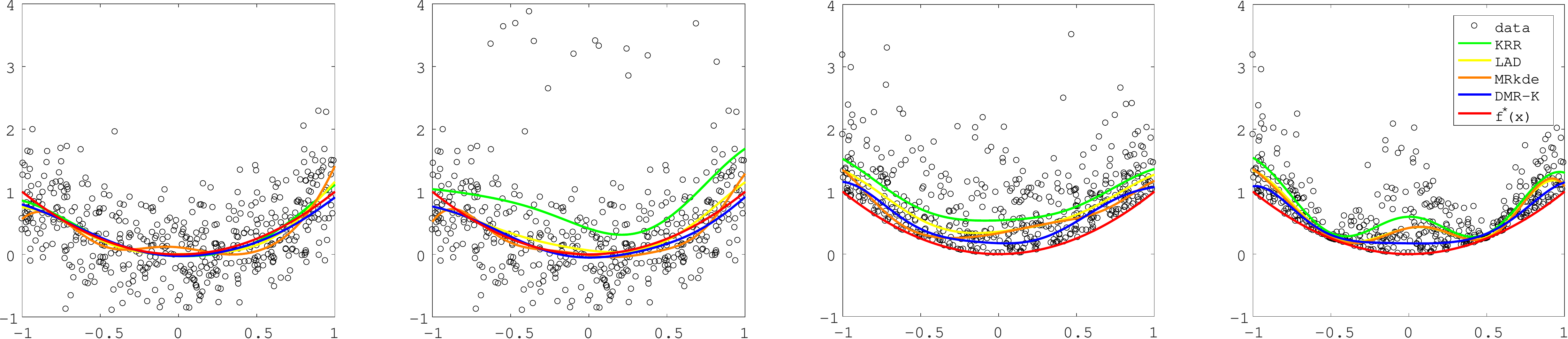}}
  \caption{\label{fig:reg-model1} Estimates of $f^*(\bm{x})$.  The top,
  middle and the bottom row are the plots when $f^{*}$ is (M1), (M2),
  and (M3), respectively. Details are given in the main text.}
 \end{figure}  

  \begin{table}[t]
   \caption{\label{reg-res1} Averages of estimation errors for (M1) over
   $30$ runs. The numbers in parentheses indicate standard deviations.
   The best and comparable methods judged by the t-test at the
   significance level 1\% are described in boldface.}
   \begin{center}
    \begin{tabular}{l|c|c|c|c}
     \hline  $\dx$ & KRR & LAD & MR$_{\KDE}$ & DMR-K \\ \hline 
     \multicolumn{5}{l}{Gauss noise}\\
     $1$ & {\bf 0.04(0.02)} & 0.06(0.02) & 0.10(0.02) & {\bf 0.05(0.03)}\\
     $5$ & {\bf 0.07(0.01)} & 0.09(0.01) & 0.19(0.02) & {\bf 0.06(0.02)}\\
     $10$ & {\bf 0.08(0.01)} & 0.11(0.02) & 0.29(0.07) & {\bf 0.08(0.04)}\\
     \multicolumn{5}{l}{Outlier noise}\\
     $1$ & 0.45(0.02) & 0.09(0.02) & 0.10(0.02) & {\bf 0.06(0.02)}\\
     $5$ & 0.44(0.02) & 0.11(0.02) & 0.20(0.02) & {\bf 0.07(0.02)}\\
     $10$ & 0.45(0.03) & 0.13(0.02) & 0.29(0.05) & {\bf 0.09(0.04)}\\
     \multicolumn{5}{l}{Skewed noise}\\
     $1$ & 0.49(0.02) & 0.35(0.02) & {\bf 0.20(0.05)} & {\bf 0.22(0.01)}\\
     $5$ & 0.50(0.02) & 0.37(0.03) & 0.28(0.02) & {\bf 0.25(0.02)}\\
     $10$ & 0.50(0.02) & 0.36(0.03) & 0.32(0.03) & {\bf 0.23(0.02)}\\
     \multicolumn{5}{l}{Nonstationary noise}\\
     $1$ & 0.31(0.02) & 0.22(0.02) & 0.17(0.03) & {\bf 0.15(0.01)}\\
     $5$ & 0.32(0.01) & 0.20(0.02) & {\bf 0.16(0.01)} & {\bf 0.15(0.01)}\\
     $10$ & 0.32(0.02) & 0.20(0.02) & 0.19(0.02) & {\bf 0.14(0.01)}\\
     \hline
    \end{tabular}
   \end{center}
  \end{table}
  \begin{table}[!t]
   \caption{\label{reg-res2} Averages of estimation errors over $30$
   runs. The left panel is for (M2), while the results for (M3) are
   shown in the right panel.}
   \begin{center}
    \begin{tabular}{l|c|c|c|c}
     \hline $\dx$ & KRR & LAD & MR$_{\KDE}$ & DMR-K \\ \hline 
     \multicolumn{5}{l}{Gauss noise}\\
     $1$ & {\bf 0.07(0.01)} & 0.08(0.01) & 0.12(0.02) & 0.08(0.03)\\
     $5$ & {\bf 0.10(0.01)} & 0.12(0.01) & 0.19(0.01) & {\bf 0.09(0.03)}\\
     $10$ & {\bf 0.10(0.01)} & 0.13(0.02) & 0.29(0.07) & {\bf 0.09(0.05)}\\
     \multicolumn{5}{l}{Outlier noise}\\
     $1$ & 0.45(0.02) & {\bf 0.10(0.02)} & 0.11(0.03) & {\bf 0.09(0.02)}\\
     $5$ & 0.44(0.02) & 0.13(0.02) & 0.20(0.02) & {\bf 0.10(0.04)}\\
     $10$ & 0.45(0.03) & 0.14(0.02) & 0.29(0.05) & {\bf 0.09(0.05)}\\
     \multicolumn{5}{l}{Skewed noise}\\
     $1$ & 0.49(0.02) & 0.35(0.02) & {\bf 0.21(0.04)} & 0.27(0.02)\\
     $5$ & 0.49(0.03) & 0.37(0.03) & 0.27(0.03) & {\bf 0.18(0.03)}\\
     $10$ & 0.49(0.02) & 0.36(0.03) & 0.33(0.04) & {\bf 0.16(0.04)}\\
     \multicolumn{5}{l}{Nonstationary noise}\\
     $1$ & 0.31(0.02) & 0.22(0.02) & {\bf 0.20(0.02)} & 0.23(0.02)\\
     $5$ & 0.31(0.02) & 0.21(0.02) & 0.15(0.01) & {\bf 0.11(0.01)}\\
     $10$ & 0.31(0.02) & 0.20(0.02) & 0.19(0.02) & {\bf 0.09(0.02)}\\
     \hline
    \end{tabular}
    \hspace{1mm}
    \begin{tabular}{l|c|c|c|c}
     \hline $\dx$ & KRR & LAD & MR$_{\KDE}$ & DMR-K \\ \hline 
     \multicolumn{5}{l}{Gauss noise}\\
     $1$ & {\bf 0.05(0.01)} & {\bf 0.06(0.02)} & 0.10(0.02) & 0.06(0.03)\\
     $5$ & {\bf 0.09(0.01)} & 0.10(0.02) & 0.19(0.01) & {\bf 0.10(0.02)}\\
     $10$ & {\bf 0.11(0.01)} & 0.13(0.01) & 0.29(0.07) & {\bf 0.10(0.03)}\\
     \multicolumn{5}{l}{Outlier noise}\\
     $1$ & 0.45(0.02) & 0.09(0.02) & 0.10(0.02) & {\bf 0.07(0.02)}\\
     $5$ & 0.44(0.03) & {\bf 0.12(0.02)} & 0.20(0.02) & {\bf 0.12(0.02)}\\
     $10$ & 0.45(0.03) & 0.15(0.02) & 0.28(0.05) & {\bf 0.11(0.04)}\\
     \multicolumn{5}{l}{Skewed noise}\\
     $1$ & 0.49(0.02) & 0.35(0.02) & {\bf 0.21(0.05)} & {\bf 0.20(0.02)}\\
     $5$ & 0.50(0.02) & 0.37(0.03) & 0.27(0.03) & {\bf 0.21(0.02)}\\
     $10$ & 0.49(0.02) & 0.36(0.03) & 0.33(0.04) & {\bf 0.17(0.03)}\\
     \multicolumn{5}{l}{Nonstationary noise}\\
     $1$ & 0.31(0.02) & 0.22(0.02) & 0.17(0.04) & {\bf 0.13(0.01)}\\
     $5$ & 0.32(0.02) & 0.20(0.02) & 0.15(0.01) & {\bf 0.12(0.01)}\\
     $10$ & 0.31(0.02) & 0.21(0.02) & 0.19(0.02) & {\bf 0.11(0.01)}\\
     \hline
    \end{tabular}
    \end{center}
  \end{table}
  Here, we investigate how DMR-K works over various noises, and compare
  it with existing modal regression methods. To estimate the conditional
  mode $f^*$, in all methods, we used the same kernel model
  $f_{\bm{\theta}}(\bm{x})=\bm{\theta}^{\top}\bkm(\bm{x})$
  in~\eqref{eqn:regression-model} and employed the Gaussian kernel where
  the width parameter was fixed at the median of the pairwise distance
  $\|\bm{x}_i-\bm{x}_j\|$ (i.e., the median trick) as done
  in~\citet{gretton2012kernel}. The following four regression methods
  were applied to the same datasets:
  \begin{itemize}
   \item \emph{Kernel ridge regression (KRR)}:
	 $f_{\bm{\theta}}(\bm{x})=\bm{\theta}^{\top}\bkm(\bm{x})$ was
	 estimated under the squared-loss with the RKHS norm
	 regularization. The regularization parameter was determined by
	 the five-hold cross-validation.

   \item \emph{Least absolute deviations (LAD)}: Absolute deviation
	 (i.e., $|y_i-f(\bm{x}_i)|$) was used as the loss function with
	 same regularization as KRR. As
	 in~\citet[Algorithm~1]{feng2017statistical}, the iteratively
	 reweighted least squares algorithm was applied to optimize the
	 parameters. The five-hold cross-validation was performed to
	 select the regularization parameter.

   \item \emph{Modal regression with kernel density estimation
	 (MR$_{\KDE}$)}: A variant of DMR-K with kernel density
	 estimation (KDE) following the naive two-step approach. As done
	 in~\citet{yao2012local}, KDE was performed to estimate the
	 joint density $p(y,\bm{x})$ where the Gauss kernel was employed
	 and the width parameters in the kernel were determined by the
	 standard least-squares
	 cross-validation~\citep{wasserman2006all}. To estimate $f^*$, a
	 similar update rule as DMR-K was derived and used similarly as
	 in Algorithm~1. Details are given in Appendix~\ref{app:mrkde}.
	 	 
   \item \emph{Direct modal regression with kernels (DMR-K)}: A proposed
	 method based on reproducing kernels.  Regarding K-LSLD, the
	 Gaussian kernel was used both for $\kx$ and $\ky$, and the
	 width parameter in each kernel is determined by the
	 leave-one-out cross-validation method in
	 Section~\ref{ssec:LOOCV}, while we fixed the regularization
	 parameter at $n^{-0.9}$ by
	 following~\citet{kanamori2012statistical}. Then,
	 $f_{\bm{\theta}}(\bm{x})$ was estimated according to
	 Algorithm~1.
  \end{itemize}
  Regarding both MR$_{\KDE}$ and DMR-K, we initialized the parameters
  $\bm{\theta}$ by LAD.
  
  We generated input data $\bm{x}_i$ from the uniform density on
  $[-1,1]^{\dx}$. Then, the output data was generated from the
  model~\eqref{regression}. For $f^*$, the following three functions
  were used:
  \begin{enumerate}[(M1)]
   \item[(M1)] $f^*(\bm{x})=\frac{1}{\dx}\sum_{j=1}^{\dx} x^{(j)}$.
       
   \item[(M2)] $f^*(\bm{x})=\sin[\frac{\pi}{\dx}\sum_{j=1}^{\dx} |x^{(j)}|]$.
	      
   \item[(M3)] $f^*(\bm{x})=\frac{1}{\dx}\sum_{j=1}^{\dx} (x^{(j)})^2$.
  \end{enumerate}
  Then, the four types of noises $\epsilon$ were generated as follows:
  \begin{itemize}
   \item \emph{Gaussian noise}: $\epsilon_i$ were sampled from the
	 Gaussian density with mean $0$ and variance $0.5$.

   \item \emph{Outlier noise}: $90\%$ of $\epsilon_i$ were sampled from
	 the Gaussian density with mean $0$ and variance $0.5$, while
	 the remainings were drawn from the uniform density on $[1,5]$
	 
   \item \emph{Skewed noise}: $\epsilon_i$ were sampled from the
	 exponential density with mean $0.5$.
	 
   \item \emph{Nonstationary-variance noise}:
	 $\epsilon_i(\bm{x}_i)=|\cos(\pi x^{(1)}_i)|\times\gamma_i$
	 where $\gamma_i$ were drawn from the exponential density with
	 mean $0.5$.
  \end{itemize}
  The total number of samples was $n=500$.  The estimation error was
  measured by
  \begin{align*}
   \frac{1}{n_{\te}}\sum_{i=1}^{n_{\te}}
   |\widehat{y}_i^{\te}-f^*(\bm{x}^{\te}_i)|,
   %\label{defi-error}
   \end{align*}
  where $n_{\te}$ denotes the number of test samples, $\bm{x}^{\te}_i$
  is a test sample generated in the same way as the training samples,
  and $\widehat{y}^{\te}_i$ is the predicted output by each method from
  $\bm{x}^{\te}_i$. We set $n_{\te}=100,000$ in this illustration.

  Fig.\ref{fig:reg-model1} plots estimates of $f^*$ by all methods in
  $\dx=1$ over all types of noises. Regarding the Gaussian noise, all
  methods give good and smooth estimates. However, regarding nonGaussian
  noises, KRR is biased from $f^*$. LAD gives a better estimate than KRR
  for the outlier noise, but does not work well to the skewed
  noise. This would be because LAD asymptotically estimates the
  conditional median, and the median is deviated from the conditional
  mode for the skewed noise.  Overall, MR$_{\KDE}$ and DMR-K perform
  well to all types of noises.
    
  Tables~\ref{reg-res1} and~\ref{reg-res2} show the averaged estimation
  errors in $\dx=1, 5, 10$. KRR achieves the best performance only for
  the Gaussian noise, while it again performs poorly to the other noises
  because the squared-loss in KRR implicitly assumes the Gaussian noise.
  LAD does not work to the skewed noise. The performance of MR$_{\KDE}$
  is good to the skewed noise only in $\dx=1$. DMR-K works the best or
  is comparable performance to the best on a wide-range of data
  dimensions and noises. Thus, our approach of directly estimating the
  gradient is promising.
  \subsection{Illustration of DMR-NN on benchmark datasets}
  Finally, we investigate the practical performance of DMR-NN on
  benchmark datasets. The datasets were downloaded from the
  web~\citep{UCIBench,CC01a}.  Each dataset was randomly divided into
  training (80\%) and test (20\%) data samples. Each data was
  standardized by the empirical means and standard deviations of the
  training samples.

  We trained a neural network $\fnn(\bm{x})$ to predict the output
  variable by least squares (LS), least absolute deviation (LAD), and
  the proposed method (DMR-NN). $\fnn(\bm{x})$ in all methods was
  modelled by a feedforward neural network with three layers where the
  numbers of hidden units were $2\dx$ and $\dx$, and the activation
  functions were all ReLU. Regarding the log-density derivative
  estimator $\rnn(y,\bm{x})$, $\mu_k^{\mathrm{NN}}(\bm{x})$ were
  modelled by a three layer neural network: The numbers of two hidden
  units were $2K$ and $K$, and the activation function was the sigmoid
  function. $\sigma_k$ were selected from $1$ to $10$ at the regular
  interval in logarithmic scale. We set $K=50$ in $\dx<30$ otherwise
  $K=100$. All parameters were optimized by Adam~\citep{kingma2015adam}
  for $500$ epochs and regularized with weight decay where the
  regularization parameter was $10^{-4}$. For DMR-NN, we performed
  pretraining for $\fnn(\bm{x})$ by LAD.
  
  For this illustration, the performance score is important. Here, we
  used the following score:
  \begin{align}
   \frac{1}{n_{\te}}\sum_{i=1}^{n_{\te}} \frac{1}{\sqrt{2\pi \sigma^2}}
   \exp\left(-\frac{(y_i^{\te}-\widehat{f}_{\mathrm{NN}}(\bm{x}^{\te}_i))^2}{2\sigma^2}\right),
   \label{emp-surrogate-gauss}
  \end{align}
  where $\sigma$ is the width parameter, $n_{\te}$ denotes the number of
  test samples, $y_i^{\te}$ and $\bm{x}^{\te}_i$ are test samples for
  input and output data respectively, and $\widehat{f}_{\mathrm{NN}}$ is
  an estimated neural network by each method. As reviewed in
  Section~\ref{ssec:review}, \eqref{emp-surrogate-gauss} is a special
  case of the surrogate empirical risk
  $\widetilde{\mathcal{R}}^{\sigma}$ (i.e.,
  $\psi(t)=\exp(-t^2/2)/\sqrt{2\pi}$ in~\eqref{surrogate}), and
  approaches to the (non-log) modal regression risk as
  $n_{\te}\rightarrow \infty$ and $\sigma\rightarrow
  0$~\citep{feng2017statistical}.  Here, we set $\sigma=n_{\te}^{-1/5}$,
  which is proved to minimize an upper bound of the excess risk in modal
  regression~\citep[Proof of Theorem~17]{feng2017statistical}.  In order
  to show that this choice of $\sigma$ is fairly good, other results on
  smaller and larger choices of $\sigma$ are presented in
  Appendix~\ref{app:bench}.
  
  The results are presented in Table~\ref{tab:bench}. Note that a larger
  number means better performance. Overall, DMR-NN works often better
  than LAD, while LS performs poorly. Thus, our method based on neural
  networks is also promising in modal regression.
  \begin{table}[t]
   \caption{\label{tab:bench} Averages of the performance
   score~\eqref{emp-surrogate-gauss} over $20$ runs. The numbers in
   parentheses indicate standard deviations.  The best and comparable
   methods judged by the t-test at the significance level 5\% are
   described in boldface. Note that larger numbers indicate better
   results.}
   \begin{center}
    \begin{tabular}{c|c|c}
     \hline  LS & LAD & DMR-NN \\ \hline 
     \multicolumn{3}{l}{space-ga ($\dx=6, n=3107$)}\\
     0.740(0.027) & {\bf 0.745(0.027)} & {\bf 0.760(0.027)}\\ \hline
     \multicolumn{3}{l}{abalone ($\dx=8, n=4177$)}\\
     0.713(0.022) & 0.739(0.020) & {\bf 0.759(0.017)}\\ \hline
     \multicolumn{3}{l}{cpusmall ($\dx=12, n=8192$)}\\
     1.476(0.010) & {\bf 1.480(0.009)} & {\bf 1.484(0.014)}\\ \hline
     \multicolumn{3}{l}{cadata ($\dx=8, n=20640$)}\\
     1.050(0.019) & 1.125(0.016) & {\bf 1.148(0.021)}\\ \hline
     \multicolumn{3}{l}{energy ($\dx=24, n=19735$)}\\
     0.956(0.039) & 1.406(0.024) & {\bf 1.484(0.015)}\\ \hline
     \multicolumn{3}{l}{superconductivty ($\dx=81, n=21263$)}\\
     1.474(0.021) & {\bf 1.541(0.016)} & 1.493(0.025)\\ \hline
     \multicolumn{3}{l}{slice loc. ($\dx=384, n=53500$)}\\
     2.504(0.006) & 2.526(0.003) & {\bf 2.537(0.001)}\\ \hline
     \multicolumn{3}{l}{sgemm ($\dx=14, n=241600$)}\\
     3.141(0.020) & 3.183(0.020) & {\bf 3.203(0.015)}\\ \hline
     \multicolumn{3}{l}{yearpred. ($\dx=90, n=515345$)}\\
     0.746(0.014) & {\bf 0.883(0.010)} & {\bf 0.886(0.026)}\\ \hline
    \end{tabular}
   \end{center}
  \end{table}
 \section{Conclusion}
 In this paper, we proposed two modal regression methods based on
 kernels and neural networks. The key idea is to directly approximate
 the gradient of the empirical modal regression risk. To this end, we
 developed direct estimators for the log-density derivative.  For the
 kernel-based modal regression method, the novel parameter update rule
 was derived based on a fixed-point method, and theoretically proved to
 have a monotonic hill-climbing property. The superior performance of
 the proposed methods was demonstrated on various artificial and
 benchmark datasets.
 \subsection*{Acknowledgement}
 The authors would like to thank Dr. Takashi Takenouchi for his helpful
 discussion.
 \appendix
 \section{Proof of Theorem~\ref{theo:analytic-solution}}
 \label{app:proof-analytic-solution}
 \begin{proof}  
  Let us denote the inner product in an RKHS by $\<\cdot,
  \cdot\>_{\rkhs}$.  Since the empirical Fisher
  divergence~\eqref{eqn:empirical-risk} can be expressed as
  \begin{align}
   \widehat{J}(r)&=\frac{1}{n}\sum_{i=1}^n\left[ 
   \frac{1}{2}\<r, k(\cdot,\bm{z}_i)\>_{\rkhs}^2 
   +\<r, \partial_{\mathrm{y}}'k(\cdot,\bm{z}_i)\>_{\rkhs}\right], 
   \label{eqn:obj-inner}
  \end{align}   
  the representer theorem for derivatives~\citep{zhou2008derivative}
  ensures that $r$ should take the following optimal form:
  \begin{align}
   r(\bm{z})=\sum_{i=1}^n\left[
   \alpha_ik(\bm{z},\bm{z}_i)+\beta_i\partial_{\mathrm{y}}'k(\bm{z},\bm{z}_i)\right], 
   \label{eqn:representer-theorem}
  \end{align}
  where $\bm{z}=(y,\bm{x})$,
  $\partial_{\mathrm{y}}'k(\bm{z},\bm{z}_i):=\parder{y'}k(\bm{z},\bm{z}')|_{\bm{z}'=\bm{z}_i}$
  with $\bm{z}'=(y',\bm{x}')$ (i.e., $\partial_{\mathrm{y}}'$ denotes
  the partial derivative with respect to the second variable of the
  kernel function $k$), and $\alpha_i$ and $\beta_i$ are coefficients to
  be estimated.  Computing the partial derivative
  of~\eqref{eqn:representer-theorem} with respect to $y$ yields
  \begin{align}
   \partial_{\mathrm{y}}\widehat{r}(\bm{z})=\sum_{i=1}^n\left[
   \alpha_i \partial_{\mathrm{y}} k(\bm{z},\bm{z}_i)
   +\beta_i\partial_{\mathrm{y}}\partial_{\mathrm{y}}'k(\bm{z},\bm{z}_i)\right],
   \label{eqn:derivative-representer-theorem}
  \end{align}
  where $\partial_{\mathrm{y}}:=\parder{y}$.

  Next, we define the $(i,j)$-th element in matrices, $\K$, $\G$ and
  $\H$, by
  \begin{align*}
   [\K]_{ij}=k(\bm{z}_i,\bm{z}_j),\ \ 
   [\G]_{ij}=\partial_{\mathrm{y}}'k(\bm{z}_i,\bm{z}_j)~~\text{and}~~
   [\H]_{ij}=\partial_{\mathrm{y}}\partial_{\mathrm{y}}'k(\bm{z}_i,\bm{z}_j).
  \end{align*}
  Then, $\bm{r}=(r(\bm{z}_1),r(\bm{z}_2),\dots,r(\bm{z}_n))^{\top}$ and
  $\partial_{\mathrm{y}}\bm{r}=(\partial_{\mathrm{y}}
  r(\bm{z}_1),\partial_{\mathrm{y}}
  r(\bm{z}_2),\dots,\partial_{\mathrm{y}} r(\bm{z}_n))^{\top}$ are
  compactly expressed as
  \begin{align}
   \bm{r}&=\K\bm{\alpha}+\G\bm{\beta}\label{eqn:vec-r}\\
   \partial_{\mathrm{y}}\bm{r}
   &=\G^{\top}\bm{\alpha}+\H\bm{\beta}.
   \label{eqn:vec-partial-r}
  \end{align}
  
  Regarding the RKHS norm,
  \begin{align}
   \|r\|_{\rkhs}^2
   &=\bm{\alpha}^{\top}\K\bm{\alpha}
   +2\sum_{i=1}^n\sum_{j=1}^n\alpha_i\beta_{j}
   \partial'_{\mathrm{y}}k(\bm{z}_i,\bm{z}_j)
   +\sum_{i=1}^n\sum_{j=1}^n\beta_i\beta_{j}
   \partial_{\mathrm{y}}\partial'_{\mathrm{y}}
   k(\bm{z}_i,\bm{z}_j)\nonumber\\\
   &=\bm{\alpha}^{\top}\K\bm{\alpha}+2\bm{\alpha}^{\top}\G\bm{\beta}
   +\bm{\beta}^{\top}\H\bm{\beta}, \label{eqn:rkhs-norm}
  \end{align}
  Substituting~\eqref{eqn:vec-r},~\eqref{eqn:vec-partial-r}
  and~\eqref{eqn:rkhs-norm} into~\eqref{eqn:obj-inner} yields
  \begin{align*}
   \widetilde{J}(r):&=\widehat{J}(r)+\frac{\lambda}{2}\|r\|^2_{\rkhs}\\
   &=\frac{1}{2n}\|\K\bm{\alpha}+\G\bm{\beta}\|^2
   +\frac{1}{n}\bm{1}_n^{\top}(\G^{\top}\bm{\alpha}+\H\bm{\beta})
   +\frac{\lambda}{2}
   (\bm{\alpha}^{\top}\K\bm{\alpha}+2\bm{\alpha}^{\top}\G\bm{\beta}
   +\bm{\beta}^{\top}\H\bm{\beta}).
  \end{align*}  
  Taking the derivatives of $\widetilde{J}$ with respect to $\bm{\alpha}$
  and $\bm{\beta}$ yields
  \begin{align*}
   \frac{\partial\widetilde{J}(r)}{\partial\bm{\alpha}}&=
   \frac{1}{n}\K(\K\bm{\alpha}+\G\bm{\beta})
   +\frac{1}{n}\G\bm{1}_n+\lambda\K\bm{\alpha}+\lambda\G\bm{\beta}\\
   &=\frac{1}{n}\K\left\{(\K+n\lambda\I_n)\bm{\alpha}+\G\bm{\beta}\right\}
   +\G\left\{\frac{1}{n}\bm{1}_n+\lambda\bm{\beta}\right\}\\
   \frac{\partial\widetilde{J}(r)}{\partial\bm{\beta}}&=
   \frac{1}{n}\G^{\top}(\K\bm{\alpha}+\G\bm{\beta})
   +\frac{1}{n}\H\bm{1}_n
   +\lambda\H\bm{\beta}+\lambda\G^{\top}\bm{\alpha}\\
   &=\frac{1}{n}\G^{\top}\left\{(\K+n\lambda\I_n)\bm{\alpha}+\G\bm{\beta}\right\}
   +\H\left\{\frac{1}{n}\bm{1}_n+\lambda\bm{\beta}\right\}.
  \end{align*}
  The optimality condition is given by 
  \begin{align*}
   (\K+n\lambda\I_n)\bm{\alpha}+\G\bm{\beta}&=\bm{0},\ \
   \frac{1}{n}\bm{1}_n+\lambda\bm{\beta}=\bm{0}.
  \end{align*}
  Thus, the optimal coefficients are given by 
  \begin{align*}
   \widehat{\bm{\alpha}}=\frac{1}{n\lambda}
   (\K+n\lambda\I_n)^{-1}\G\bm{1}_n, \ \ 
   \widehat{\bm{\beta}}=-\frac{1}{n\lambda}\bm{1}_n.
  \end{align*}
  Substituting $\widehat{\bm{\alpha}}$ and $\widehat{\bm{\beta}}$
  into~\eqref{eqn:representer-theorem} completes the proof.
 \end{proof}  
 \section{Details for Leave-One-Out Cross-Validation}
 \label{app:LOOCV}
 Here, we show that the LOOCV score can be efficiently computed by
 following~\citet{kanamori2012statistical}. The notations in
 Section~\ref{app:proof-analytic-solution} are inherited

 Let us denote the collection of data samples except $\bm{z}_l$ by
 $\D_l~(i.e., \D\setminus\bm{z}_l)$. K-LSLD from $\D_l$ is given by
 \begin{align*}
  \widehat{r}^{(l)}(\bm{z})=\sum_{\substack{i=1\\ i\neq l}}^n
  \left[\widehat{\alpha}^{(l)}_ik(\bm{z},\bm{z}_i)
  +\widehat{\beta}^{(l)}_i\partial_{\mathrm{y}}'k(\bm{z},\bm{z}_i)\right],
 \end{align*}
 where    
 \begin{align*}
  \widehat{\bm{\alpha}}^{(l)}&=\frac{1}{(n-1)\lambda}
  (\bm{K}^{(l)}+(n-1)\lambda\I_{n-1})^{-1}\bm{G}^{(l)}\bm{1}_{n-1}, \ \
  \widehat{\bm{\beta}}^{(l)}=-\frac{1}{(n-1)\lambda}\bm{1}_{n-1}.
 \end{align*}
 In the equations above, $\bm{K}^{(l)}$ and $\bm{G}^{(l)}$ are $\bm{K}$
 and $\bm{G}$ except $\bm{z}_l$, respectively. Then, the LOOCV score
 can be computed as
 \begin{align*}
  \LOOCV=\frac{1}{n}\sum_{l=1}^n\left[\frac{1}{2}
  \{\widehat{r}^{(l)}(\bm{z}_l)\}^2
  +\partial_{\mathrm{y}}\widehat{r}^{(l)}(\bm{z}_l)\right].
 \end{align*}
 However, to naively compute the LOOCV score, we need to compute the
 inverse of $n-1$ by $n-1$ matrix for each $\widehat{\bm{a}}^{(l)}$,
 which is time-consuming.
 
 To cope with this problem, we derive an equivalent form of
 $\widehat{r}^{(l)}$. $\widehat{\bm{\alpha}}^{(l)}$ can be regarded as
 the solution of the optimization problem,
 \begin{align*}
  \widehat{\bm{\alpha}}^{(l)}=\argmin_{\bm{\alpha}\in\R{n-1}}\left[
  \frac{1}{2}\bm{\alpha}^{\top}(\bm{K}^{(l)}+(n-1)\lambda\I_{n-1})\bm{\alpha}
  -\frac{1}{(n-1)\lambda}\bm{1}_{n-1}^{\top}
  \bm{G}^{(l)}\bm{\alpha}\right].
 \end{align*}
 Here, we solve an alternative optimization problem as
 \begin{align}
  \widetilde{\bm{\alpha}}^{(l)}:=\argmin_{\bm{\alpha}\in\R{n}}\left[
  \frac{1}{2}\bm{\alpha}^{\top}(\bm{K}+(n-1)\lambda\I_{n})\bm{\alpha}
  -\frac{1}{(n-1)\lambda}(\bm{1}_{n}-\bm{e}_l)^{\top}
  \bm{G}\bm{\alpha}\right]\quad\mathrm{s.t.}\quad\widetilde{\alpha}^{(l)}_l=0,
  \label{tilde-alpha}
 \end{align}
 where $\bm{e}_l$ is the unit vector with the $l$-th element being $1$.
 With $\widetilde{\bm{\alpha}}^{(l)}$, $\widehat{r}^{(l)}$ can
 be equivalently expressed as
 \begin{align}
  \widehat{r}^{(l)}(\bm{z})=\sum_{i=1}^n
  \left[\widetilde{\alpha}^{(l)}_ik(\bm{z},\bm{z}_i)
  +\widetilde{\beta}^{(l)}_i\partial_{\mathrm{y}}' k(\bm{z},\bm{z}_i)\right], 
 \end{align}
 where
 \begin{align*}
  \widetilde{\bm{\beta}}^{(l)}
  :=-\frac{1}{(n-1)\lambda}(\bm{1}_{n}-\bm{e}_l).
 \end{align*}  
 Applying the method of Lagrange multipliers to~\eqref{tilde-alpha}
 yields
 \begin{align}
  \widetilde{\bm{\alpha}}^{(l)}=(\bm{K}+(n-1)\lambda\I_{n})^{-1}
  \left\{\frac{1}{(n-1)\lambda}\bm{G}(\bm{1}_{n-1}-\bm{e}_l)
  +t_l\bm{e}_l\right\},   
 \end{align}
 where $t_l$ is set such that $\widetilde{\alpha}^{(l)}_l=0$. The key
 point is that unlike $\widehat{\bm{\alpha}}^{(l)}$, computing the
 inverse of the $n$ by $n$ matrix only once is sufficient to obtain all
 $\widetilde{\bm{\alpha}}^{(l)}$.
 
 Next, we derive the analytic form of the LOOCV score. Let
 \begin{align*}
  \bm{A}=(\widetilde{\bm{\alpha}}^{(1)},\dots,\widetilde{\bm{\alpha}}^{(n)})\ \ 
  \mathrm{and}\ \ 
  \bm{B}=(\widetilde{\bm{\beta}}^{(1)},\dots,\widetilde{\bm{\beta}}^{(n)}).
 \end{align*}
 Then,
 \begin{align}
  \bm{A}=\bm{L}(\bm{S}-\bm{T})\ \ \mathrm{and}\ \
  \bm{B}=-\frac{1}{(n-1)\lambda}\bm{E},
 \end{align}
 where $\bm{L}:=(\bm{K}+(n-1)\lambda\I_{n})^{-1}$,
 $\bm{S}:=\frac{1}{(n-1)\lambda}\bm{G}\bm{E}$, 
 \begin{align*}
  [\bm{E}]_{ij}:=
  \begin{cases}
   0 & i=j,\\
   1 & i\neq j,
  \end{cases}
  \ \ \mathrm{and}\ \ 
  [\bm{T}]_{ij}:=
  \begin{cases}
   [\bm{L}\bm{S}]_{ii}/[\bm{L}]_{ii} & i=j,\\
   0 & i\neq j.
  \end{cases}   
 \end{align*}
 Finally, the LOOCV score can be computed analytically as
 \begin{align}
  \LOOCV=\frac{1}{n}\left\{\frac{1}{2}\widetilde{\bm{r}}^{\top}\widetilde{\bm{r}}
  +\bm{1}_n^{\top}\partial_{\mathrm{y}}\widetilde{\bm{r}}\right\},
 \end{align}
 where 
 \begin{align*}
  \widetilde{\bm{r}}&=
  (\widehat{r}^{(1)}(\bm{z}_1),\widehat{r}^{(2)}(\bm{z}_2),
  \dots,\widehat{r}^{(n)}(\bm{z}_n))^{\top}=
  (\K\odot\bm{\A}^{\top}+\G\odot\bm{B}^{\top})\bm{1}_n\\
  \partial_{\mathrm{y}}\widetilde{\bm{r}}&=
  (\partial_{\mathrm{y}}\widehat{r}^{(1)}(\bm{z}_1),
  \partial_{\mathrm{y}}\widehat{r}^{(2)}(\bm{z}_2),\dots,
  \partial_{\mathrm{y}}\widehat{r}^{(n)}(\bm{z}_n))^{\top}
  =(\G^{\top}\odot\bm{A}^{\top} +\H\odot\bm{B}^{\top})\bm{1}_n.
 \end{align*}
 The symbol $\odot$ denotes element-wise multiplication.
 \section{Proof of Theorem~\ref{theo:LSUR-rule}}
 \label{app:LSUR-rule}
 With the assumption that $\widehat{\alpha}_l=0$ for all $l$, we compute
 $\widehat{D}[\bm{\theta}_2|\bm{\theta}_1]$ as
 \begin{align}
  \widehat{D}[\bm{\theta}_2|\bm{\theta}_1]&=
  \frac{1}{n}\sum_{i=1}^n\int_0^1
  \widehat{r}(\bm{\theta}(t)^{\top}\bkm(\bm{x}_i),\bm{x}_i)
  \bkm(\bm{x}_i)^{\top}(\bm{\theta}_2-\bm{\theta}_1)\intd t
  \nonumber\\&=\frac{1}{n}\sum_{i,l=1}^n\underbrace{\left[\int_0^1
  \frac{y_l-\bm{\theta}(t)^{\top}\bkm(\bm{x}_i)}
  {n\lambda\sigma_{\mathrm{y}}^2}
  \varphi\left\{\frac{(\bm{\theta}(t)^{\top}\bkm(\bm{x}_i)-y_l)^2}
  {2\sigma_{\mathrm{y}}^2}\right\}
  \bkm(\bm{x}_i)^{\top}(\bm{\theta}_2-\bm{\theta}_1)\intd t\right]}_{(\star)}
  \kx(\bm{x}_i,\bm{x}_l),
  \label{app:path-integral-esitmate}
 \end{align}
 By the substitution
 $Y_l=\frac{y_l-\bm{\theta}^{\top}(t)\bkm(\bm{x}_i)}{\sigma_{\mathrm{y}}}$,
 the integral $(\star)$ is computed as
 \begin{align}
  (\star)&=-\frac{1}{n\lambda} \int _{Y_l^{(1)}}^{Y_l^{(2)}}
  Y_l\varphi\left(\frac{Y_l^2}{2}\right)\intd Y_l =\frac{1}{n\lambda}
 \left[\phi\left\{\frac{(\bm{\theta}_2^{\top}\bm{k}(\bm{x}_i)-y_l)^2}
  {2\sigma_{\mathrm{y}}^2}\right\}
  -\phi\left\{\frac{(\bm{\theta}_1^{\top}\bm{k}(\bm{x}_i)-y_l)^2}
  {2\sigma_{\mathrm{y}}^2}\right\}\right], \label{app:integral-star}
 \end{align}
 where we used $\frac{\intd Y_l}{\intd t}
 =\frac{(\bm{\theta}_2-\bm{\theta}_1)^{\top}\bkm(\bm{x}_i)}
 {\sigma_{\mathrm{y}}}$ from~\eqref{linear-path},
 $Y_l^{(1)}=\frac{y_l-\bm{\theta}_1^{\top}\bkm(\bm{x}_i)}{\sigma_{\mathrm{y}}}$
 and
 $Y_l^{(2)}=\frac{y_l-\bm{\theta}_2^{\top}\bkm(\bm{x}_i)}{\sigma_{\mathrm{y}}}$.
 
 Then, substituting~\eqref{app:integral-star}
 into~\eqref{app:path-integral-esitmate} yields
 \begin{align*}
  \widehat{D}[\bm{\theta}_2|\bm{\theta}_1]&=
  \frac{1}{n^2\lambda}\sum_{i,l=1}^n
  \kx(\bm{x}_i,\bm{x}_l)
  \left[\phi\left\{\frac{(\bm{\theta}_2^{\top}\bkm(\bm{x}_i)-y_l)^2}
  {2\sigma_{\mathrm{y}}^2}\right\}
  -\phi\left\{\frac{(\bm{\theta}_1^{\top}\bkm(\bm{x}_i)-y_l)^2}
  {2\sigma_{\mathrm{y}}^2}\right\}\right]\\
  &\geq\frac{1}{n^2\lambda}\sum_{i,l=1}^n \kx(\bm{x}_i,\bm{x}_l)
  \varphi\left\{\frac{(\bm{\theta}_1^{\top}\bkm(\bm{x}_i)-y_l)^2}
  {2\sigma_{\mathrm{y}}^2}\right\}
  \left\{\frac{(\bm{\theta}_1^{\top}\bkm(\bm{x}_i)-y_l)^2}
  {2\sigma_{\mathrm{y}}^2}
  -\frac{(\bm{\theta}_2^{\top}\bkm(\bm{x}_i)-y_l)^2}
  {2\sigma_{\mathrm{y}}^2}\right\}\\
  &=\frac{1}{2}
  \left\{\bm{\theta}_1^{\top}\bm{H}(\bm{\theta}_1)\bm{\theta}_1
  -\bm{\theta}_2^{\top}\bm{H}(\bm{\theta}_1)\bm{\theta}_2
  -2(\bm{\theta}_1-\bm{\theta}_2)^{\top}\bm{h}(\bm{\theta}_1)\right\},
 \end{align*}
 where we applied a well-known inequality for convex functions as
 \begin{align*}
  \phi(t_2)-\phi(t_1)\geq \varphi(t_1)(t_1-t_2),
 \end{align*}
 where $\varphi(t):=-\frac{\intd}{\intd t}\phi(t)$.
 
 By $\bm{\theta}_1\leftarrow\bm{\theta}^{\tau}$ and
 $\bm{\theta}_2\leftarrow\bm{\theta}^{\tau+1}$, we have
 \begin{align*}
  \widehat{D}[\bm{\theta}^{\tau+1}|\bm{\theta}^{\tau}]&\geq 
  \frac{1}{2}
  \left\{\bm{\theta}^{\tau\top}\bm{H}(\bm{\theta}^{\tau})\bm{\theta}^{\tau}
  -\bm{\theta}^{\tau+1\top}\bm{H}(\bm{\theta}^{\tau})\bm{\theta}^{\tau+1}
  -2(\bm{\theta}^{\tau}-\bm{\theta}^{\tau+1})^{\top}\bm{h}(\bm{\theta}^{\tau})
  \right\}\\  
  &=\frac{1}{2}
  \left\{\bm{\theta}^{\tau\top}\bm{H}(\bm{\theta}^{\tau})\bm{\theta}^{\tau}
  -\bm{\theta}^{\tau+1\top}\bm{H}(\bm{\theta}^{\tau})\bm{\theta}^{\tau+1}
  -2(\bm{\theta}^{\tau}-\bm{\theta}^{\tau+1})^{\top}\bm{H}(\bm{\theta}^{\tau})
  \bm{\theta}^{\tau+1}\right\}\\ 
  &=\frac{1}{2}
  \left\{\bm{\theta}^{\tau\top}\bm{H}(\bm{\theta}^{\tau})\bm{\theta}^{\tau}
  +\bm{\theta}^{\tau+1\top}\bm{H}(\bm{\theta}^{\tau})\bm{\theta}^{\tau+1}
  -2\bm{\theta}^{\tau\top}\bm{H}(\bm{\theta}^{\tau})
  \bm{\theta}^{\tau+1}\right\}\\ 
  &=\frac{1}{2}
  (\bm{\theta}^{\tau}-\bm{\theta}^{\tau+1})^{\top}
  \bm{H}(\bm{\theta}^{\tau})(\bm{\theta}^{\tau}-\bm{\theta}^{\tau+1}),
 \end{align*}
 where we used the relation
 $\bm{h}(\bm{\theta}^{\tau})=\bm{H}(\bm{\theta}^{\tau})\bm{\theta}^{\tau+1}$
 in~\eqref{fixed-point1} on the first line.  Since $\bm{H}(\bm{\theta})$
 is assumed to be positive definite, the right-hand side is positive for
 $\bm{\theta}^{\tau}\neq\bm{\theta}^{\tau+1}$. Thus, the proof is
 completed.
 \section{Details of MR$_{\KDE}$}
 \label{app:mrkde}
 \subsection{Risk with the joint probability density function}
 Since the conditional and joint densities yield the same maximizer with
 respect to the output variable, the conditional mode function
 $f_{\mathrm{M}}$ can be defined from the joint density $p(y,\bm{x})$ as
 \begin{align}
  f_{\mathrm{M}}(\bm{x}):=\argmax_t p(t|\bm{x})
  =\argmax_t p(t,\bm{x}).
  \label{defi-joint-mode}
 \end{align}
 Thus, the following risk alternative to the modal regression risk can
 be used for conditional mode estimation:
 \begin{align*}
  \mathcal{R}_{\mathrm{J}}(f):=\int p(f(\bm{x}),\bm{x})
  p(\bm{x})\intd\bm{x}.
 \end{align*}
 The following inequality, which follows from~\eqref{defi-joint-mode},
 ensures that the maximizer of $\mathcal{R}_{\mathrm{J}}(f)$ is
 $f_{\mathrm{M}}$:
 \begin{align*}
  \mathcal{R}_{\mathrm{J}}(f)\leq \int p(f_{\mathrm{M}}(\bm{x}),\bm{x})
  p(\bm{x})\intd\bm{x}.
 \end{align*}
 With a parametrized model $f_{\bm{\theta}}(\bm{x})$ as in the kernel
 model, the empirical version of $\mathcal{R}_{\mathrm{J}}$ can be
 obtained as
 \begin{align*}
  \widehat{\mathcal{R}}_{\mathrm{J}}(\bm{\theta}):=
  \frac{1}{n}\sum_{i=1}^n p(f_{\bm{\theta}}(\bm{x}_i),\bm{x}_i).
 \end{align*}
 In practice, we need to estimate the joint density $p(y, \bm{x})$ to
 approximate $\widehat{\mathcal{R}}_{\mathrm{J}}(\bm{\theta})$. Below,
 we employ kernel density estimation (KDE) for the joint density $p(y,
 \bm{x})$ as done in~\citet{yao2012local} and derive an update rule
 similar as DMR-K.
 \subsection{Update rule based on a fixed-point method}
 Let us define KDE with the Gaussian kernel to the joint density
 $p(y,\bm{x})$ by
 \begin{align*}
  \widehat{p}_{\KDE}(y,\bm{x})
  =\frac{1}{nZ}\sum_{l=1}^n 
  \exp\left(-\frac{(y-y_l)^2}{2h_\mathrm{y}^2}\right)
  \exp\left(-\frac{\|\bm{x}-\bm{x}_l\|^2}{2h_\mathrm{x}^2}\right),
 \end{align*}
 where $Z=(2\pi)^{(\dx+1)/2}h_\mathrm{y}h_\mathrm{x}^{\dx}$, and
 $h_\mathrm{y}$ and $h_\mathrm{x}$ are positive width parameters.  Then,
 $\widehat{p}_{\KDE}(y,\bm{x})$ enables us to approximate
 $\widehat{\mathcal{R}}_{\mathrm{J}}(\bm{\theta})$ as 
 \begin{align*}
  \widetilde{\mathcal{R}}_{\KDE}(\bm{\theta}):= \frac{1}{n}\sum_{i=1}^n
  \widehat{p}_{\KDE}(f_{\bm{\theta}}(\bm{x}_i),\bm{x}_i).
 \end{align*}
 Computing the gradient of $\widetilde{\mathcal{R}}_{\KDE}(\bm{\theta})$
 with respect to $\bm{\theta}$ yields
 \begin{align*}
  \parder{\bm{\theta}}\widetilde{\mathcal{R}}_{\KDE}(\bm{\theta})
  =\frac{1}{n}\sum_{i=1}^n\parder{\bm{\theta}} f_{\bm{\theta}}(\bm{x}_i)
  \parder{y}\widehat{p}_{\KDE}(y,\bm{x})&=
  \frac{1}{n^2h_\mathrm{y}^2Z}\left\{
  \bm{h}_{\KDE}(\bm{\theta})-\H_{\KDE}(\bm{\theta})\bm{\theta}\right\},
 \end{align*} 
 where $f_{\bm{\theta}}(\bm{x})=\bm{\theta}^{\top}\bkm(\bm{x})$,
 \begin{align*}
  \parder{y}\widehat{p}_{\KDE}(y,\bm{x})&=\frac{1}{n Z}
  \sum_{l=1}^n\frac{y_l-y}{h_\mathrm{y}^2}
  \exp\left(-\frac{(y-y_l)^2}{2h_\mathrm{y}^2}\right)
  \exp\left(-\frac{\|\bm{x}-\bm{x}_l\|^2}{2h_\mathrm{x}^2}\right)\\
  \H_{\KDE}(\bm{\theta})&=\sum_{i=1}^n\sum_{l=1}^n
  \exp\left(-\frac{(\bm{\theta}^{\top}\bkm(\bm{x}_i)-y_l)^2}
  {2h^2_\mathrm{y}}\right)
  \exp\left(-\frac{\|\bm{x}_i-\bm{x}_l\|^2}{2h^2_\mathrm{x}}\right)
  \bkm(\bm{x}_i)\bkm(\bm{x}_i)^{\top},\\
  \bm{h}_{\KDE}(\bm{\theta})&=\sum_{i=1}^n\sum_{l=1}^n
  y_l \exp\left(-\frac{(\bm{\theta}^{\top}\bkm(\bm{x}_i)-y_l)^2}
  {2h^2_{\mathrm{y}}}\right)
  \exp\left(-\frac{\|\bm{x}_i-\bm{x}_l\|^2}{2h^2_\mathrm{x}}\right)
  \bkm(\bm{x}_i).	  
 \end{align*}
 Setting the right-hand side above to equal to zero leads to the
 following update rule:
 \begin{align}
  \bm{\theta}\leftarrow
  \H^{-1}_{\KDE}(\bm{\theta})\bm{h}_{\KDE}(\bm{\theta}).
  \label{fixed-point-KDE}
 \end{align}
 Eq.\eqref{fixed-point-KDE} is iteratively used to update $\bm{\theta}$
 as in Algorithm~1.
 \section{Validity of the performance score~\eqref{emp-surrogate-gauss}}
 \label{app:bench}
 \citet{feng2017statistical} discussed that the meaning of the maximizer
 of the surrogate risk $\widetilde{\mathcal{R}}^{\sigma}$, which
 includes our performance score~\eqref{emp-surrogate-gauss} as a special
 case, is different depending on the width parameter $\sigma$: When
 $\sigma$ approaches zero, the maximizer is asymptotically a conditional
 \emph{mode} estimator. On the other hand, the maximizer is a
 (robustified) conditional \emph{mean} estimator as $\sigma,
 n\rightarrow\infty$~\citep[Table~2]{feng2017statistical}. In accord
 with the theory, the right panel in Table~\ref{tab:bench-others} shows
 LS and LAD outperform DMR-NN for large $\sigma$ because these methods
 estimate the conditional mean and median asymptotically, while DMR-NN
 often works better than LS and LAD when $\sigma$ is small (Left panel
 in Table~\ref{tab:bench-others}, $\sigma=0.01$).  Our choice of
 $\sigma=n_{\te}^{-1/5}=(0.2n)^{-1/5}$ in the main text\footnote{Let us
 remind that we used 20\% of data samples for test in experiments on
 benchmark datasets (i.e., $n_{\te}=0.2n$).} is in fact a middle of
 these two panels in Table~\ref{tab:bench-others} and approximately
 $0.09\leq \sigma \leq 0.27$ among all datasets. Thus, it seems to be a
 fairly good choice because the standard deviations in the left panel of
 Table~\ref{tab:bench-others} are often large and the result for too
 small $\sigma$ could be unreliable.
 \begin{table}[h]
  \caption{\label{tab:bench-others} Averages of the performance
  score~\eqref{emp-surrogate-gauss} over $20$ runs when $\sigma=0.01$
  (left panel) and $\sigma=1.0$ (right panel). The numbers in
  parentheses indicate standard deviations.  The best and comparable
  methods judged by the t-test at the significance level 5\% are
  described in boldface. Note that larger numbers indicate better
  results.}
  \begin{center}
   \begin{tabular}{c|c|c}
    \hline LS & LAD & DMR-NN \\ \hline \multicolumn{3}{l}{space-ga
    ($\dx=6, n=3107$)}\\ 0.876(0.170) & 0.913(0.228) & {\bf
    1.075(0.211)}\\ \hline \multicolumn{3}{l}{abalone ($\dx=8,
    n=4177$)}\\ {\bf 0.873(0.156)} & {\bf 0.911(0.182)} & {\bf
    0.884(0.187)}\\ \hline \multicolumn{3}{l}{cpusmall ($\dx=12,
    n=8192$)}\\ 3.519(0.263) & {\bf 4.007(0.270)} & 3.642(0.477)\\
    \hline \multicolumn{3}{l}{cadata ($\dx=8, n=20640$)}\\ 1.332(0.099)
    & 1.575(0.101) & {\bf 1.661(0.144)}\\ \hline
    \multicolumn{3}{l}{energy ($\dx=24, n=19735$)}\\ 1.152(0.126) &
    2.534(0.163) & {\bf 2.872(0.130)}\\ \hline
    \multicolumn{3}{l}{superconductivty ($\dx=81, n=21263$)}\\
    3.204(0.252) & {\bf 5.077(0.198)} & {\bf 5.048(0.500)}\\ \hline
    \multicolumn{3}{l}{slice loc. ($\dx=384, n=53500$)}\\ 14.325(0.897)
    & 20.394(0.906) & {\bf 24.868(1.000)}\\ \hline
    \multicolumn{3}{l}{sgemm ($\dx=14, n=241600$)}\\ 10.416(0.860) &
    {\bf 14.305(0.899)} & 12.785(1.031)\\ \hline
    \multicolumn{3}{l}{yearpred. ($\dx=90, n=515345$)}\\ 0.763(0.023) &
    {\bf 0.928(0.021)} & {\bf 0.914(0.082)}\\ \hline
   \end{tabular}
  %\end{center}
 \hspace{5mm}
 \begin{tabular}{c|c|c}
  \hline  LS & LAD & DMR-NN \\ \hline 
  \multicolumn{3}{l}{space-ga ($\dx=6, n=3107$)}\\
  {\bf 0.357(0.003)} & {\bf 0.357(0.003)} & {\bf 0.358(0.003)}\\ \hline
  \multicolumn{3}{l}{abalone ($\dx=8, n=4177$)}\\
  {\bf 0.344(0.003)} & {\bf 0.345(0.003)} & 0.342(0.003)\\ \hline
  \multicolumn{3}{l}{cpusmall ($\dx=12, n=8192$)}\\
  {\bf 0.394(0.000)} & 0.394(0.000) & {\bf 0.394(0.000)}\\ \hline
  \multicolumn{3}{l}{cadata ($\dx=8, n=20640$)}\\
  0.368(0.001) & {\bf 0.369(0.001)} & 0.367(0.001)\\ \hline
  \multicolumn{3}{l}{energy ($\dx=24, n=19735$)}\\
  0.350(0.003) & {\bf 0.363(0.002)} & 0.359(0.002)\\ \hline
  \multicolumn{3}{l}{superconductivty ($\dx=81, n=21263$)}\\
  {\bf 0.384(0.001)} & {\bf 0.384(0.001)} & 0.374(0.005)\\ \hline
  \multicolumn{3}{l}{slice loc. ($\dx=384, n=53500$)}\\
  0.399(0.000) & 0.399(0.000) & {\bf 0.399(0.000)}\\ \hline
  \multicolumn{3}{l}{sgemm ($\dx=14, n=241600$)}\\
  {\bf 0.398(0.000)} & 0.398(0.000) & 0.398(0.000)\\ \hline
  \multicolumn{3}{l}{yearpred. ($\dx=90, n=515345$)}\\
  0.331(0.001) & {\bf 0.336(0.000)} & 0.311(0.021)\\ \hline
 \end{tabular}
 \end{center}
 \end{table}

 \bibliography{../../../papers}
 \bibliographystyle{abbrvnat}

\end{document}